\documentclass[runningheads,a4paper]{llncs}
\pdfoutput=1
\usepackage{amssymb}
\usepackage{graphicx}
\usepackage{amsmath}
\usepackage{comment}
\usepackage{multirow}
\usepackage{color}

\newcommand {\commenting}[1]{\textcolor[rgb]{1,0,0} {#1}}

\usepackage{array}
\newcolumntype{L}[1]{>{\raggedright\let\newline\\\arraybackslash\hspace{0pt}}m{#1}}
\newcolumntype{C}[1]{>{\centering\let\newline\\\arraybackslash\hspace{0pt}}m{#1}}

\usepackage{url}
\urldef{\mailsa}\path|{W.Pei-1, H.Dibeklioglu, D.M.J.Tax, L.J.P.vanderMaaten}@tudelft.nl|
\urldef{\mailsc}\path|L.J.P.vanderMaaten}@springer.com|

\begin{document}

\mainmatter  % start of an individual contribution

\title{Time Series Classification using the Hidden-Unit Logistic Model}

\titlerunning{Time Series Classification using the Hidden-Unit Logistic Model}

\author{Wenjie Pei%
%\thanks{Please note that the LNCS Editorial assumes that all authors have used
%the western naming convention, with given names preceding surnames. This determines
%the structure of the names in the running heads and the author index.}%
\and Hamdi Dibeklio\u{g}lu \and David M.J. Tax \and Laurens van der Maaten}
\authorrunning{Time Series Classification using the Hidden-Unit Logistic Model}

\institute{Pattern Recognition \& Bioinformatics Group, Delft University of Technology\\
Mekelweg 4, Delft, The Netherlands\\
\mailsa\\
%\url{http://www.springer.com/lncs}
}


\toctitle{Lecture Notes in Computer Science}
\tocauthor{Authors' Instructions}
\maketitle

\begin{abstract}
% abstract
We present a new model for time series classification, called
the hidden-unit logistic model, that uses binary stochastic
hidden units to model latent structure in the data. The hidden
units are connected in a chain structure that models temporal
dependencies in the data. Compared to the prior models for time
series classification such as the hidden conditional random
field, our model can model very complex decision boundaries
because the number of latent states grows exponentially with
the number of hidden units. We demonstrate the strong
performance of our model in experiments on a variety of
(computer vision) tasks, including handwritten character
recognition, speech recognition, facial expression, and action
recognition. We also present a state-of-the-art system for
facial action unit detection based on the hidden-unit logistic
model.

\end{abstract}

%%%%%%%%% BODY TEXT
\section{Introduction}

Time series classification is the problem of assigning a single
label to a sequence of observations (i.e., to a time series).
Time series classification has a wide range of applications in
computer vision. A state-of-the-art model for time series
classification problem is the hidden-state conditional random
field (HCRF)~\cite{HCRF}, which models latent structure in the
data using a chain of $k$-nomial latent variables. The HCRF has
been successfully used in, amongst others, gesture
recognition~\cite{HCRF_gesture}, object
recognition~\cite{HCRF}, and action
recognition~\cite{HCRF_action}. An important limitation of the
HCRF is that the number of model parameters grows linearly with
the number of latent states in the model. This implies that the
training of complex models with a large number of latent states
is very prone to overfitting, whilst models with smaller
numbers of parameters may be too simple to represent a good
classification function. In this paper, we propose to
circumvent this problem of the HCRF by replacing each of the
$k$-nomial latent variables by a collection of $H$ binary
stochastic hidden units. To keep inference tractable, the
hidden-unit chains are conditionally independent given the time
series and the label. Similar ideas have been explored before
in discriminative RBMs~\cite{RBMs} for standard classification
problems and in hidden-unit CRFs \cite{HuCRFs} for sequence
labeling. The binary stochastic hidden units allow the
resulting model, which we call the hidden-unit logistic model
(HULM), to represent $2^H$ latent states using only $O(H)$
parameters. This substantially reduces the amount of data
needed to successfully train models without overfitting, whilst
maintaining the ability to learn complex models with
exponentially many latent states. Exact inference in our
proposed model is tractable, which makes parameter learning via
(stochastic) gradient descent very efficient. We show the
merits of our hidden-unit logistic model in experiments on
computer-vision tasks ranging from online character recognition
to activity recognition and facial expression analysis.
Moreover, we present a system for facial action unit detection
that, with the help of the hidden-unit logistic model, achieves
state-of-the-art performance on a commonly used benchmark for
facial analysis.

The remainder of this paper is organized as follows. Section 2 reviews prior work on time series classification. Section 3 introduces our hidden-unit logistic model and describes how inference and learning can be performed in the model. In Section 4, we present the results of experiments comparing the performance of our model with that of state-of-the-art time series classification models on a range of classification tasks. In Section 5, we present a new state-of-the-art system for facial action unit detection based on the hidden-unit logistic model. Section 6 concludes the paper.% and discusses potential directions for future work.

%------------------------------------------------------------------------
\section{Related Work}

There is a substantial amount of prior work on time series classification. Much of this work is based on the use of (kernels based on) dynamic time warping (\emph{e.g.}, \cite{jeni2014}) or on hidden Markov models (HMMs)~\cite{HMM_model}. The HMM is a generative model that models the time series data in a chain of latent $k$-nomial features. Class-conditional HMMs are commonly combined with class priors via Bayes' rule to obtain a time series classification models. Alternatively, HMMs are also frequently used as the base model for Fisher kernel~\cite{fisher_kernel}, which constructs a time series representation that consists of the gradient a particular time series induces in the parameters of the HMM; the resulting representations can be used on standard classifiers such as SVMs. Some recent work has also tried to learn the parameters of the HMM in such a way as to learn Fisher kernel representations that are well-suited for nearest-neighbor classification \cite{FKL}. HMMs have also been used as the base model for probability product kernels \cite{Jebara04}, which fit a single HMM on each time series and define the similarity between two time series as the inner product between the corresponding HMM distributions. A potential drawback of these approaches is that they perform classification based on (rather simple) generative models of the data that may not be well suited for the discriminative task at hand. By contrast, we opt for a discriminative model that does not waste model capacity on features that are irrelevant for classification.

In contrast to HMMs, conditional random fields (CRFs;~\cite{CRFs}) are discriminative models that are commonly used for sequence labeling of time series using so-called linear-chain CRFs. Whilst standard linear-chain CRFs achieve strong performance on very high-dimensional data (e.g., in natural language processing), the linear nature of most CRF models limits their ability to learn complex decision boundaries. Several sequence labeling models have been proposed to address this limitation, amongst which are latent-dynamic CRFs \cite{LDCRF}, conditional neural fields \cite{peng_nips09}, and hidden-unit CRFs~\cite{HuCRFs}. These models introduce stochastic or deterministic hidden units that model latent structure in the data, allowing these models to represent nonlinear decision boundaries. As these prior models were designed for sequence labeling (assigning a label to each frame in the time series), they cannot readily be used for time series classification (assigning a single label to the entire time series). Our hidden-unit logistic model may be viewed as an adaptation of sequence labeling models with hidden units to the time series classification problem. As such, it is closely related to the hidden CRF model \cite{HCRF}. The key difference between our hidden-unit logistic model and the hidden CRF is that our model uses a collection of binary stochastic hidden units instead of a single $k$-nomial hidden unit, which allows our model to represent exponentially more states with the same number of parameters.

An alternative approach to expanding the number of hidden states of the HCRF is the infinite HCRF (iHCRF), which employs a Dirichlet process to determine the number of hidden states. Inference in the iHCRF can be performed via collapsed Gibbs sampling~\cite{iHCRF} or variational inference~\cite{HCRF-DPM}. Whilst theoretically facilitating infinitely many states, the modeling power of the iHCRF is, however, limited to the number of ``represented'' hidden states. Unlike our model, the number of parameters in the iHCRF thus still grows linearly with the number of hidden states.

\section{Hidden-Unit Logistic Model}
The hidden-unit logistic model is a probabilistic graphical model that receives a time series as input, and is trained to produce a single output label for this time series. Like the hidden-state CRF, the model contains a chain of hidden units that aim to model latent temporal features in the data, and that form the basis for the final classification decision. The key difference with the HCRF is that the latent features are model in $H$ binary stochastic hidden units, much like in a (discriminative) RBM. These hidden units $\mathbf{z}_t$ can model very rich latent structure in the data: one may think about them as carving up the data space into $2^H$ small clusters, all of which may be associated with particular clusters. The parameters of the temporal chains that connect the hidden units may be used to differentiate between features that are ``constant'' (i.e., that are likely to be presented for prolonged lengths of time) or that are ``volatile'' (i.e., that tend to rapidly appear and disappear). Because the hidden-unit chains are conditionally independent given the time series and the label, they can be integrated out analytically when performing inference or learning.

Suppose we are given a time series $\mathbf{x}_{1, \dots, T} = \{\mathbf{x}_1, \dots, \mathbf{x}_T \}$ of length $T$ in which the observation at the $t$-th time step is denoted by $\mathbf{x}_t \in \mathbb{R}^D$. Conditioned on this time series, the hidden-unit logistic model outputs a distribution over vectors $\mathbf{y}$ that represent the predicted label using a $1$-of-$K$ encoding scheme (i.e., a one-hot encoding): $\forall k: y_k \in \{0, 1\}$ and $\sum_{k} y_k = 1$.

%\begin{figure}[t]
%\begin{center}
%%\fbox{\rule{0pt}{2in} \rule{0.9\linewidth}{0pt}}
%\includegraphics[width=0.85\linewidth]{image/hidden-unit_logistic_model}
%\end{center}
%\caption{Factor graph of the hidden-unit logistic model. }
%\label{fig:logistic_model}
%\end{figure}

\begin{figure}[t]
\begin{center}
   \includegraphics[width=0.7\linewidth]{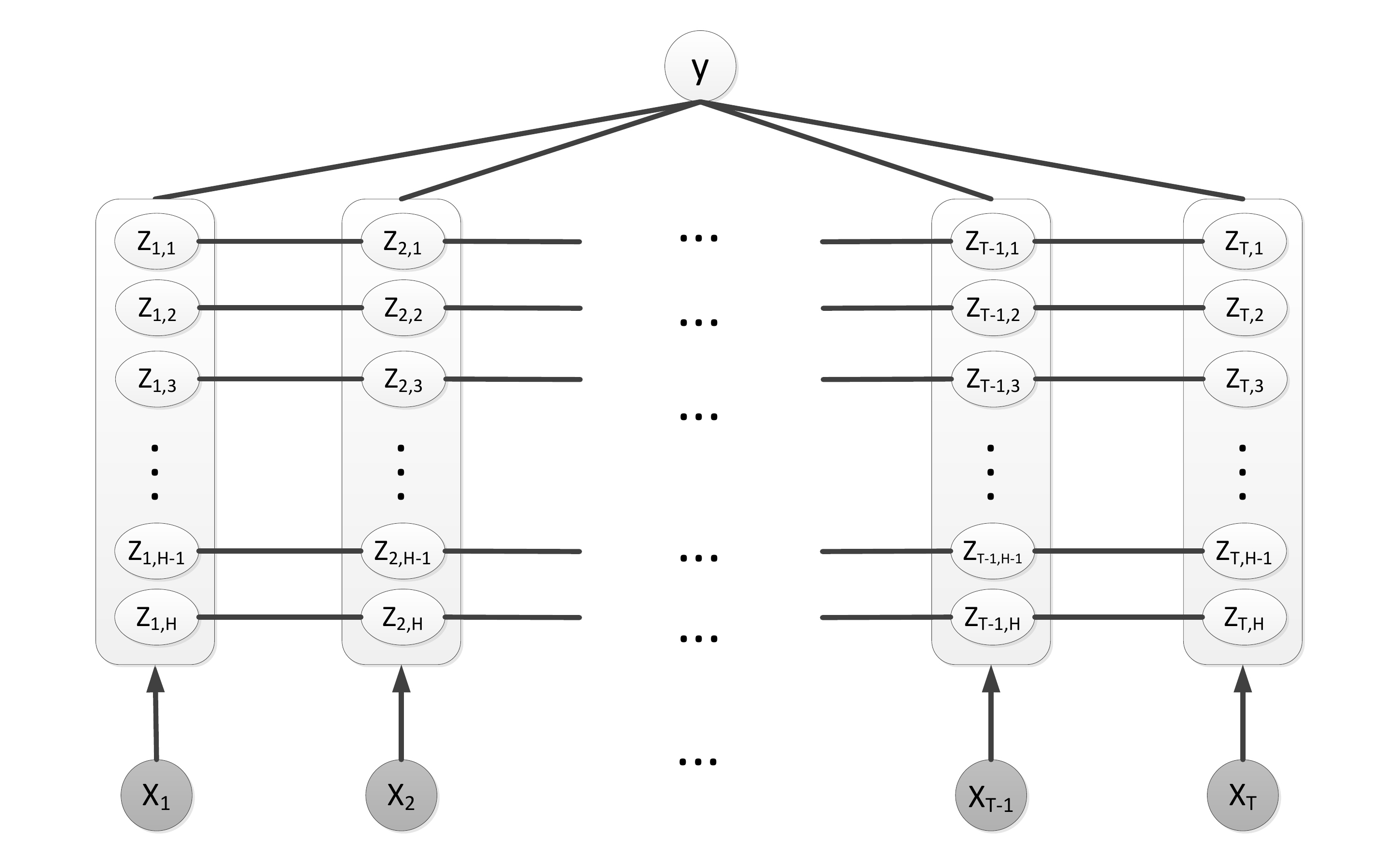}
\end{center}
\vspace{-0.2in}
   \caption{Graphical model of the hidden-unit logistic model.}
\label{fig:logistic_model}
\end{figure}

Denoting the stochastic hidden units at time step $t$ by $\mathbf{z}_t \in \{0, 1\}^H$, the hidden-unit logistic model defines the conditional distribution over label vectors using a Gibbs distribution in which all hidden units are integrated out:
\begin{equation}
p(\mathbf{y} | \mathbf{x}_{1, \dots, T}) = \frac{\sum_{\mathbf{z}_{1, \dots, T}} \mathrm{exp}\{E(\mathbf{x}_{1, \dots, T}, \mathbf{z}_{1, \dots, T}, \mathbf{y} )\}}{Z(\mathbf{x}_{1, \dots, T})}.
\label{eqn:first}
\end{equation}
Herein, $Z(\mathbf{x}_{1, \dots, T})$ denotes a partition function that normalizes the distribution, and is given by:
\begin{equation}
Z(\mathbf{x}_{1,\dots, T}) = \sum_{\mathbf{y'}}{ \sum_{\mathbf{z}_{1, \dots, T}'} {\mathrm{exp}\{ E(\mathbf{x}_{1, \dots, T}, \mathbf{z}_{1, \dots, T}', \mathbf{y}' )\}}}.
\label{eqn:partition}
\end{equation}
The energy function of the hidden-unit logistic model is defined as:
\begin{equation}
%\begin{split}
%&E(\mathbf{x}_{1, \dots, T}, \mathbf{z}_{1, \dots, T}, \mathbf{y} )  =  \mathbf{z}_1^\top \boldsymbol{\pi} + \mathbf{z}_T^\top \boldsymbol{\tau} + \mathbf{c}^\top \mathbf{y} +\nonumber\\
%&~~~~~\sum_{t=2}^{T}\mathbf{z}_{t-1}^T\mathrm{diag}(\mathbf{A})\mathbf{z}_{t} + \sum_{t=1}^\top \left[ \mathbf{z}_t^\top \mathbf{W} \mathbf{x}_t + \mathbf{z}_t^\top \mathbf{V} \mathbf{y} +  \mathbf{z}^\top_t \mathbf{b} \right].
% \label{eqn:energy1}
% \end{split}
\resizebox{1.0\linewidth}{!}{
$E(\mathbf{x}_{1, \dots, T}, \mathbf{z}_{1, \dots, T}, \mathbf{y} )  =  \mathbf{z}_1^\top \boldsymbol{\pi} + \mathbf{z}_T^\top \boldsymbol{\tau} + \mathbf{c}^\top \mathbf{y} + \sum_{t=2}^{T}\mathbf{z}_{t-1}^T\mathrm{diag}(\mathbf{A})\mathbf{z}_{t} + \sum_{t=1}^T \left[ \mathbf{z}_t^\top \mathbf{W} \mathbf{x}_t + \mathbf{z}_t^\top \mathbf{V} \mathbf{y} +  \mathbf{z}^\top_t \mathbf{b} \right].$
}
\label{eqn:energy1}
\end{equation}
The graphical model of the hidden-unit logistic model is shown in Figure~\ref{fig:logistic_model}.

%Next to a number of bias terms, the energy function in Equation~\ref{eqn:energy1} consists of three main components: (1) a term with parameters $\mathbf{W}$ that measures to what extent particular latent features are present in the data; (2) a term parametrized by $\mathbf{A}$ that measures the compatibility between corresponding hidden units at time step $t-1$ and $t$; and (3) a prediction term with parameters $\mathbf{V}$ that measures the compatibility between the latent features $\mathbf{z}_{1,\dots,T}$ and the label vector $\mathbf{y}$. Please note that hidden units in consecutive time steps are connected using a chain structure rather than fully connected; we opt for this structure because exact inference is intractable when consecutive hidden units are fully connected. Intuitively, the hidden-unit logistic model thus assigns a high probability to a label (for a particular input) when there are hidden unit states that are both ``compatible" with the observed data and with a particular label. As the hidden units can take $2^H$ different states, this leads to a model that can represent highly nonlinear decision boundaries. Below, we describe how to perform inference and learning in the hidden-unit logistic model.

\subsection{Inference}
\vspace{-0.02in}
%The main inferential problem given an observation $\mathbf{x}$ is the calculation of most likely label, which can be solved by
%\begin{eqnarray}
%\mathbf{y}^* = \underset{\mathbf{y}}{\mathrm{argmax}}\{{p(\mathbf{y}|\mathbf{x}_{1,\dots, T})}\}
%\end{eqnarray}

The main inferential problem given an observation $\mathbf{x}_{1,\dots,T}$ is the evaluation of predictive distribution $p(\mathbf{y} | \mathbf{x}_{1, \dots, T})$. The key difficulty in computing this predictive distribution is the sum over all $2^{H \times T}$ hidden unit states:
\begin{equation}
M(\mathbf{x}_{1,\dots, T}, \mathbf{y}) = \sum_{\mathbf{z}_{1, \dots, T}} \mathrm{exp}\{E(\mathbf{x}_{1, \dots, T}, \mathbf{z}_{1, \dots, T}, \mathbf{y} )\}.
\end{equation}
The chain structure of the hidden-unit logistic model allows us to employ a standard forward-backward algorithm that can compute $M(\cdot)$ in computational time linear in $T$.
%\begin{flalign}
%& E(\mathbf{x}_{1, \dots, T}, \mathbf{z}_{1, \dots, T}, \mathbf{y} )  =  \mathbf{z}_1^\top \mathbf{\pi} + \mathbf{z}_T^\top \mathbf{\tau} + \mathbf{c}^\top \mathbf{y} +  &\nonumber\\
%& \sum_{t=1}^{T}{[\sum_{h=1}^H ({z}_{t-1, h} A_h {z}_{t, h}) + \mathbf{z}_t^\top \mathbf{W} \mathbf{x}_t + \mathbf{z}_t^\top \mathbf{V} \mathbf{y} +  \mathbf{z}^\top_t \mathbf{b}]}.&
% \label{eqn:energy2}
%\end{flalign}

%\begin{eqnarray}
% E(\mathbf{x}_{1, \dots, T}, \mathbf{z}_{1, \dots, T}, \mathbf{y} )  =  \mathbf{z}_1^\top \mathbf{\pi} + \mathbf{z}_T^\top \mathbf{\tau} + \mathbf{c}^\top \mathbf{y} +  \nonumber\\
% \sum_{t=1}^{T}{\sum_{h=1}^H \left({z}_{t-1, h} A_h {z}_{t, h} + z_{t,h} \mathbf{W}_h   \mathbf{x}_t + z_{t,h} \mathbf{V}_h \mathbf{y} + z_{t,h} b_h\right)}.
% \label{eqn:energy2}
%\end{eqnarray}
Specifically, defining potential functions that contain all terms that involve time $t$ and hidden unit $h$:
%\begin{flalign}
% &\Psi_t(\mathbf{x_t}, \mathbf{z_{t-1}}, \mathbf{z_t}, \mathbf{y})&
%   \nonumber \\
%&  =     \mathrm{exp}\{ \sum_{h=1}^H (\mathbf{z}_{t-1, h} \mathbf{A}_{h} \mathbf{z}_{t, h}) + \mathbf{z}_{t}^\top \mathbf{W}   \mathbf{x}_t + \mathbf{z}_t^\top \mathbf{V} y + \mathbf{z}_t^\top \mathbf{b}_h\} & \nonumber\\
%  &=  \prod_{h=1}^H (\mathrm{exp}\{  \mathbf{z}_{t-1, h} \mathbf{A}_{h} z_{t, h} + z_{t,h} \mathbf{W}_h   \mathbf{x}_t + z_{t,h} \mathbf{V}_h y + z_{t,h} b_h\}) & \nonumber \\
%  & = \prod_{h=1}^H {\Psi_{t,h}(\mathbf{x_t}, z_{t-1,h}, z_{t,h}, \mathbf{y})} &
%\end{flalign}
%\begin{flalign}
%& \Psi_{t,h}(\mathbf{x}_t, z_{t-1,h}, z_{t,h}, \mathbf{y}) \nonumber \\
%& = \mathrm{exp}\{  \mathbf{z}_{t-1, h} \mathbf{A}_{h} z_{t, h} + z_{t,h} \mathbf{W}_h   \mathbf{x}_t + z_{t,h} \mathbf{V}_h \mathbf{y} + z_{t,h} b_h\}, \nonumber
%\end{flalign}
\begin{equation}
\Psi_{t,h}(\mathbf{x}_t, z_{t-1,h}, z_{t,h}, \mathbf{y})
= \mathrm{exp}\{  \mathbf{z}_{t-1, h} \mathbf{A}_{h} z_{t, h} + z_{t,h} \mathbf{W}_h   \mathbf{x}_t + z_{t,h} \mathbf{V}_h \mathbf{y} + z_{t,h} b_h\}, \nonumber
\end{equation}
%\begin{eqnarray}
% & \Psi_{t,h}(\mathbf{x_t}, z_{t-1,h}, z_{t,h}, \mathbf{y}) \nonumber \\
% = & \mathrm{exp}\{  \mathbf{z}_{t-1, h} \mathbf{A}_{h} z_{t, h} + z_{t,h} \mathbf{W}_h   \mathbf{x}_t + z_{t,h} \mathbf{V}_h \mathbf{y} + z_{t,h} b_h\}
%\end{eqnarray}
ignoring bias terms, and introducing virtual hidden units $\mathbf{z}_0 = \mathbf{0}$ at time $t=0$, we can rewrite $M(\cdot)$ as:
%\begin{align}
%M(\cdot) & = \sum_{\mathbf{z}_{1, \dots, T}}  {\prod_{t=1}^{T} \prod_{h=1}^H {\Psi_{t,h}(\mathbf{x_t}, z_{t-1,h}, z_{t,h}, \mathbf{y})} } \nonumber \\
%& = \prod_{h=1}^H \left[ \sum_{z_{1,h}, \dots, z_{T,h}} {\prod_{t=1}^{T} {\Psi_{t,h}(\mathbf{x}_t, z_{t-1,h}, z_{t,h}, \mathbf{y})} } \right] \nonumber\\
%& =  \prod_{h=1}^H  \left[  \sum_{z_{T-1,h}} {\Psi_{T,h}(\mathbf{x}_{T}, z_{T-1,h}, z_{T,h}, \mathbf{y})}  \right. \nonumber \\
% & \left. ~~~~~~~~\sum_{z_{T-2,h}} {\Psi_{T-1,h}(\mathbf{x}_{T-1}, z_{T-2,h}, z_{T-1,h}, \mathbf{y})}  \dots  \right]\nonumber.
% \label{eqn:sum_over}
%\end{align}

\noindent\resizebox{\linewidth}{!}{
\begin{minipage}{1.1\linewidth}
\begin{align}
M(\cdot) & = \sum_{\mathbf{z}_{1, \dots, T}}  {\prod_{t=1}^{T} \prod_{h=1}^H {\Psi_{t,h}(\mathbf{x_t}, z_{t-1,h}, z_{t,h}, \mathbf{y})} }  = \prod_{h=1}^H \left[ \sum_{z_{1,h}, \dots, z_{T,h}} {\prod_{t=1}^{T} {\Psi_{t,h}(\mathbf{x}_t, z_{t-1,h}, z_{t,h}, \mathbf{y})} } \right] \nonumber\\
& =  \prod_{h=1}^H  \left[  \sum_{z_{T-1,h}} {\Psi_{T,h}(\mathbf{x}_{T}, z_{T-1,h}, z_{T,h}, \mathbf{y})} \sum_{z_{T-2,h}} {\Psi_{T-1,h}(\mathbf{x}_{T-1}, z_{T-2,h}, z_{T-1,h}, \mathbf{y})}  \dots  \right]\nonumber.
 \label{eqn:sum_over}
\end{align}
\end{minipage}
}
In the above derivation, it should be noted that the product over hidden units $h$ can be pulled outside the sum over all states $\mathbf{z}_{1, \dots, T}$ because the hidden-unit chains are conditionally independent given the data $\mathbf{x}_{1, \dots, T}$ and the label $\mathbf{y}$. Subsequently, the product over time $t$ can be pulled outside the sum because of the (first-order) Markovian chain structure of the temporal connections between hidden units.

%\begin{eqnarray}
%%&& \sum_{\mathbf{z}_{1, \dots, T}} \mathrm{exp}\{E(\mathbf{x}_{1, \dots, T}, \mathbf{z}_{1, \dots, T}, \mathbf{y} )\} \nonumber \\
% = &\sum_{\mathbf{z}_{1, \dots, T}}  {\prod_{t=1}^{T} \prod_{h=1}^H {\Psi_{t,h}(\mathbf{x_t}, z_{t-1,h}, z_{t,h}, \mathbf{y})} } \nonumber\\
% = &\prod_{h=1}^H \{ \sum_{z_{(1,h), \dots, (T,h)}}
% {\prod_{t=1}^{T} {\Psi_{t,h}(\mathbf{x_t}, z_{t-1,h}, z_{t,h}, \mathbf{y})} } \} \nonumber\\
% = & \prod_{h=1}^H  \{  \sum_{z_{T-1,h}} {{\Psi_{T,h}(\mathbf{x}_{T}, z_{T-1,h}, z_{T,h}, \mathbf{y})} } \nonumber \\
%& \sum_{z_{T-2,h}} {{\Psi_{T-1,h}(\mathbf{x}_{T-1}, z_{T-2,h}, z_{T-1,h}, \mathbf{y})} }
% \dots \}
% \label{eqn:sum_over}
%\end{eqnarray}

In particular, the required quantities can be evaluated using the forward-backward algorithm, in which we define the forward messages $\alpha_{t, h, k}$ with $k \in \{0, 1\}$ as:
\[ \alpha_{t, h, k} = \sum_{z_{1,h}, \dots, z_{t-1, h}}\prod_{t'=1}^{t} {\Psi_{t',h}(\mathbf{x}_{t'}, z_{t'-1,h}, z_{t',h}=k, \mathbf{y})},\]
and the backward messages $\beta_{t, h, k}$ as:
\begin{equation*}
 \beta_{t, h, k} = \sum_{z_{t+1,h}, \dots, z_{T, h}} \prod_{t'=t+1}^{T} {\Psi_{t',h}(\mathbf{x}_{t'+1}, z_{t',h}=k, z_{t'+1,h}, \mathbf{y})}.
\end{equation*}

These messages can be calculated recursively as follows:
%\setlength\arraycolsep{1pt}
%\resizebox{\linewidth}{!}{
%\begin{minipage}{1.05\linewidth}
%\begin{equation*}
%  \alpha_{t,h,k}  =  \sum_{i \in \{0, 1\} }{\Psi_{t,h}(\mathbf{x}_t, z_{t-1,h}=i, z_{t,h}=k, \mathbf{y} )} \alpha_{t-1,h,i}
%\end{equation*}
%\end{minipage}
%}
%\setlength\arraycolsep{1pt}
%\resizebox{\linewidth}{!}{
%\begin{minipage}{1.05\linewidth}
%\begin{equation*}
%\beta_{t,h,k} = \sum_{i \in \{0, 1\}}{\Psi_{t+1,h}(\mathbf{x}_{t+1}, z_{t,h}=k, z_{t+1,h}=i, \mathbf{y} )} \beta_{t+1, h,i}
%\end{equation*}
%\end{minipage}
%}
\setlength\arraycolsep{1pt}
\begin{flalign}
 & \alpha_{t,h,k}  =  \sum_{i \in \{0, 1\} }{\Psi_{t,h}(\mathbf{x}_t, z_{t-1,h}=i, z_{t,h}=k, \mathbf{y} )} \alpha_{t-1,h,i} \nonumber \\
\setlength\arraycolsep{1pt}
& \beta_{t,h,k} = \sum_{i \in \{0, 1\}}{\Psi_{t+1,h}(\mathbf{x}_{t+1}, z_{t,h}=k, z_{t+1,h}=i, \mathbf{y} )} \beta_{t+1, h,i}\nonumber.
 \label{eqn:sum_product}
\end{flalign}

%\setlength\arraycolsep{1pt}
%\begin{eqnarray}
%  \alpha_{t,h,k} & = & \sum_{i \in \{0,1\} }{\Psi_{t,h}(\mathbf{x}_t, \mathbf{z}_{t-1,h}=i, \mathbf{z}_{t,h}=k, \mathbf{y} )} \alpha_{t-1,h,i}\nonumber \\
% \alpha_{1,h,k}  & = & {\Psi_{1,h}(\mathbf{x}_{1}, \mathbf{z}_{0}, z_{1,h}=k, \mathbf{y})}  \mathrm{exp}\{ z_{1,h}^\top \mathbf{\pi_h}\}  \nonumber\\
% \alpha_{T,h,k} & = & \sum_{i \in \{0,1\} } \mathrm{exp}\{ z_{T,h} \mathbf{\tau}_h\} \nonumber\\
% && {\Psi_{T,h}(\mathbf{x}_T, z_{T-1,h}=i, z_{T,h}=k, \mathbf{y} )} \alpha_{T-1,h, i}
% \label{eqn:alpha}
%\end{eqnarray}
%and
%\setlength\arraycolsep{1pt}
%\begin{eqnarray}
% \beta_{t,h,k} &=& \sum_{i \in \{1,0\}}{\Psi_{t+1,h}(\mathbf{x}_{t+1}, \mathbf{z}_{t,h}=k, \mathbf{z}_{t+1,h}=i, \mathbf{y} )} \beta_{t+1, h,i} \nonumber\\
% \beta_{1, h, k} &=& {\Psi_{1,h}(\mathbf{x}_{1}, z_{0,h}, z_{1,h}, \mathbf{y})}  \mathrm{exp}\{ z_{1,h}^\top \mathbf{\pi_h}\} \cdot \beta_{1, h, k} \nonumber\\
% \beta_{T, h, k} &=& \mathrm{exp}\{ z_{T,h}^\top \mathbf{\pi_h}\}
% \label{eqn:beta}
%\end{eqnarray}
The value of $M(\mathbf{x}_{1,\dots, T}, \mathbf{y})$ can readily be computed from the resulting forward messages or backward messages:
\begin{eqnarray}
 M(\mathbf{x}_{1,\dots, T}, \mathbf{y})
=  \prod_{h=1}^H \left( \sum_{k \in \{0,1\}}\alpha_{T,h,k}\right)
= \prod_{h=1}^H \left( \sum_{k \in \{0,1\}}\beta_{1,h,k} \right).
\end{eqnarray}
To complete the evaluation of the predictive distribution, we compute the partition function of the predictive distribution by summing $M(\mathbf{x}_{1,\dots, T}, \mathbf{y})$ over all $K$ possible labels: $Z(\mathbf{x}_{1,\dots, T}) = \sum_{\mathbf{y'}}  M(\mathbf{x}_{1,\dots, T}, \mathbf{y'})$. Indeed, inference in the hidden-unit logistic model is linear in both the length of the time series $T$ and in the number of hidden units $H$.\\

Another inferential problem that needs to be solved during parameter learning is the evaluation of the marginal distribution over a chain edge:
\[
\xi_{t,h,k,l} = P(z_{t,h}= k, z_{t+1, h}=l | \mathbf{x}_{1,\dots, T}, \mathbf{y}	).
\]
Using a similar derivation, it can be shown that this quantity can also be computed from the forward and backward messages:
%\noindent\resizebox{\linewidth}{!}{
%\begin{minipage}{1.1\linewidth}
%\begin{equation*}
%\xi_{t,h,k,l} = \frac{ \alpha_{t,h,k} \cdot \Psi_{t+1, h}(\mathbf{x}_{t+1}, z_{t,h}=k, z_{t+1,h}=l,y) \cdot \beta_{t+1, h,l}}{\sum_{k \in \{0,1\}}\alpha_{T,h,k}}.
%\end{equation*}
%\end{minipage}
%}
\begin{equation*}
\xi_{t,h,k,l} = \frac{ \alpha_{t,h,k} \cdot \Psi_{t+1, h}(\mathbf{x}_{t+1}, z_{t,h}=k, z_{t+1,h}=l,y) \cdot \beta_{t+1, h,l}}{\sum_{k \in \{0,1\}}\alpha_{T,h,k}}.
\end{equation*}
\vspace{-0.1in}

\subsection{Parameter Learning}
\vspace{-0.02in}
Given a training set $\mathcal{D} = \{ (\mathbf{x}^(n)_{1, \dots, T}, \mathbf{y}^(n) ) \}_{n = 1, \dots, N}$ containing $N$ pairs of time series and their associated label. We learn the parameters $\boldsymbol{\Theta} = \{\pi, \tau, \mathbf{A}, \mathbf{W}, \mathbf{V}, \mathbf{b}, \mathbf{c}\}$ of the hidden-unit logistic model by maximizing the conditional log-likelihood of the training data with respect to the parameters:
\begin{align}
\mathcal{L}(\Theta) = & \sum_{n=1}^N \log p \left(\mathbf{y}^{(n)} | \mathbf{x}^{(n)}_{1,\dots, T}\right)& \nonumber\\
= & \sum_{n=1}^N \left[ \log M \left(\mathbf{x}^{(n)}_{1,\dots, T}, \mathbf{y}^{(n)}\right) - \log \sum_{\mathbf{y'}}  M \left(\mathbf{x}^{(n)}_{1,\dots, T}, \mathbf{y'}\right)\right].
\label{eqn:L}
\end{align}
We augment the conditional log-likelihood with
L2-regularization terms on the parameters $\mathbf{A}$,
$\mathbf{W}$, and $\mathbf{V}$. As the objective function is
not amenable to closed-form optimization (in fact, it is not
even a convex function), we perform optimization using
stochastic gradient descent on the negative conditional
log-likelihood. The gradient of the conditional log-likelihood
with respect to a parameter $\theta \in \Theta$ is given by:
%\begin{align}
%\frac{\partial \mathcal{L}}{\partial \theta} = &~\mathbb{E} \left[\frac{\partial{E(\mathbf{x}_{1, \dots, T}, \mathbf{z}_{1, \dots, T}, \mathbf{y} )}}{\partial{\theta}} \right]_{P(\mathbf{z}_{1,\dots, T} | \mathbf{x}_{1,\dots, T}, \mathbf{y})}\nonumber\\
%- &~\mathbb{E} \left[\frac{\partial{E(\mathbf{x}_{1, \dots, T}, \mathbf{z}_{1, \dots, T}, \mathbf{y} )}}{\partial{\theta}} \right]_{P(\mathbf{z}_{1,\dots, T}, \mathbf{y} | \mathbf{x}_{1,\dots, T})},
%\end{align}
\begin{equation}
\resizebox{\linewidth}{!}{
$\frac{\partial \mathcal{L}}{\partial \theta} = \mathbb{E} \left[\frac{\partial{E(\mathbf{x}_{1, \dots, T}, \mathbf{z}_{1, \dots, T}, \mathbf{y} )}}{\partial{\theta}} \right]_{P(\mathbf{z}_{1,\dots, T} | \mathbf{x}_{1,\dots, T}, \mathbf{y})} - \mathbb{E} \left[\frac{\partial{E(\mathbf{x}_{1, \dots, T}, \mathbf{z}_{1, \dots, T}, \mathbf{y} )}}{\partial{\theta}} \right]_{P(\mathbf{z}_{1,\dots, T}, \mathbf{y} | \mathbf{x}_{1,\dots, T})},$
}\end{equation}
where we omitted the sum over training examples for brevity.
The required expectations can readily be computed using the
inference algorithm described in the previous subsection.

For example, defining $r(\Theta) = z_{t-1, h} \mathbf{A}_{h} z_{t, h}+ z_{t,h} \mathbf{W}_h \mathbf{x}_t + z_{t,h} \mathbf{V}_h y + z_{t,h} b_h$ for notational simplicity, the first expectation can be computed as follows:
\begin{align}
 &\mathbb{E}\left[\frac{\partial{E(\mathbf{x}_{1, \dots, T}, \mathbf{z}_{1, \dots, T}, \mathbf{y} )}}{\partial{\theta}}\right]_{P(\mathbf{z}_{1,\dots, T} | \mathbf{x}_{1,\dots, T}, \mathbf{y})}& \nonumber\\
%& = \sum_{\mathbf{z_{1,\dots, T}}}
%P(\mathbf{z}_{1,\dots, T} | \mathbf{x}_{1,\dots, T}, \mathbf{y})
%\frac{\partial{E(\mathbf{x}_{1, \dots, T}, \mathbf{z}_{1, \dots, T}, \mathbf{y} )}}{\partial{\theta}} & \nonumber\\
& = \sum_{\mathbf{z_{1, \dots, T}}} P(\mathbf{z}_{1,\dots, T} | \mathbf{x}_{1,\dots, T}, \mathbf{y})
\left(\sum_{t=1}^{T} \sum_{h=1}^{H}
\frac{\partial{r(\Theta)}}{\partial{\theta}}\right) & \nonumber \\
& = \sum_{t=1}^{T} \sum_{k \in \{0,1\}} \sum_{l \in \{0,1\}}
\left( \xi_{t-1,h,k,l} \cdot \frac{\partial{r(\Theta)}}{\partial{\theta}} \right). &
\end{align}
The second expectation is simply an average of these expectations over all $K$ possible labels $\mathbf{y}$.

%\begin{displaymath}
%\xi_{t,h,k,l} = \frac{ \mathrm{exp}\{ \mathbf{c}^\top \mathbf{y} \}  \cdot V}{M}
%\end{displaymath}
%where
%\begin{displaymath}
%\setlength\arraycolsep{1pt}
%\begin{array}{ll}
%V & =
%\left( \alpha_{t,h,k} \cdot \Psi_{t+1, h}(\mathbf{x}_{t+1}, \mathbf{z}_{t,h}=k, \mathbf{z}_{t+1,h}=l,y) \cdot \beta_{t+1, h,l}\right) \\
% &\prod_{h^{'}=1}^{h-1} (\sum_{k \in \{0,1\}}\alpha_{T,h^{'},k})
%\prod_{h^{'}=h+1}^H (\sum_{k \in \{0,1\}}\alpha_{T,h^{'},k})
%\end{array}
%\end{displaymath}
%In this way, we can calculate the result of $\frac{\partial{\log{ M(\mathbf{x}_{1,\dots, T}, \mathbf{y})}}}{\partial{\boldsymbol{\theta}}}$ and thereby obtain the gradients of $\mathcal{L}$ with respect to model parameters $\boldsymbol{\theta}$.
%It should be noted that the negative conditional log-likelihood of the hidden-unit Logistic model is a non-convex function. Hence, we can only converge to a local maximum of the conditional log-likelihood with the stochastic gradient descent(SGD) optimizer.

%------------------------------------------------------------------------
\section{Experiments}
To evaluate the performance of the hidden-unit logistic model,
we conducted classification experiments on five different
problems involving time series features: (1) an online
handwritten character data set (OHC) \cite{character}; (2) a
data set of Arabic spoken digits (ASD) \cite{Arabic}; (3) the
Cohn-Kanade extended facial expression data set (CK+)
\cite{CK}; (4) the MSR Action 3D data set (Action)
\cite{Action3D}; and (5) the MSR Daily Activity 3D data set
(Activity) \cite{MSRActivity}. The five data sets are
introduced in \ref{data_set}, the experimental setup is
presented in~\ref{setup}, and the results of the experiments
are in \ref{result}.
% We compared the performance of our hidden-unit logistic model with that of several state-of-the-art time series classification algorithms.

\vspace{-0.05in}
\subsection{Data Sets}
\label{data_set}
\vspace{-0.02in}
The online handwritten character dataset~\cite{character} is a
pen-trajectory time series data set that consists of three
dimensions at each time step, \emph{viz.}, the pen movement in
the $x$-direction and $y$-direction, and the pen pressure. The
data set contains 2858 time series with an average length of
120 frames. Each time series corresponds to a single
handwritten character that has one of 20 labels. We pre-process
the data by windowing the features of 10 frames into a single
feature vector with $30$ dimensions.
%(This also reduces the length of the time series by a factor of ten.)

The Arabic spoken digit dataset contains 8800
utterances~\cite{Arabic}, which were collected by asking 88
Arabic native speakers to utter all 10 digits ten times. Each
time series consists of 13-dimensional MFCCs which were sampled
at 11,025Hz, 16-bits using a Hamming window. We enrich the
features by windowing 3 frames into 1 frames resulting in the
$13 \times 3$ dimensions for each frame of the features while
keeping the same length of time series.

The Cohn-Kanade extended facial expression data set~\cite{CK}
contains 593 image sequences (videos) from 123 subjects. Each
video shows a single facial expression. The videos have an
average length of 18 frames. A subset of 327 of the videos,
which have validated label corresponding to one of seven
emotions (anger, contempt, disgust, fear, happiness, sadness,
and surprise), are used in our experiments. We adopt the
publicly available shape features used
in~\cite{maaten2012action} as the feature representation for
our experiments. These features represent each frame by the
variation of 68 feature point locations ($x, y$) with respect
to the first frame \cite{CK}, which leads to 136-dimensional
feature representation for each frame in the video.

The MSR Action 3D data set~\cite{Action3D} consists of RGB-D
videos of people performing certain actions. The data set
contains 567 videos with an average length of 41 frames. Each
video should be classified into one of 20 actions such as
``high arm wave'', ``horizontal arm wave'', and ``hammer''. We
use the real-time skeleton tracking algorithm
of~\cite{skeleton} to extract the 3D joint positions from the
depth sequences. We use the 3D joint position features
(pairwise relative positions) proposed in~\cite{MSRActivity} as
the feature representation for the frames in the videos. Since
we track a total of 20 joints, the dimensionality of the
resulting feature representation is $3 \times {20 \choose 2} =
570$, where ${20 \choose 2}$ is the number of pairwise
distances between joints and 3 is dimensionality of the $(x, y,
z)$ coordinate vectors.
%\comment{(Comment: to Laurens: the performance of our model using HON4D feature is bad: $63\%$ accuracy versus $76\%$ with FFT + SVM )}

The MSR Daily Activity 3D data set~\cite{MSRActivity} contains
RGB-D videos of people performing daily activities. The data
set also contains 3D skeletal joint positions, which are
extracted using the Kinect SDK. The videos need to be
classified into one of 16 activity types, which include
``drinking'', ``eating'', ``reading book'', \emph{etc.} Each
activity is performed by 10 subjects in two different poses
(namely, while sitting on a sofa and while standing), which
leads to a total of 320 videos. The videos have an average
length of 193 frames. To represent each frame, we extract
$570$-dimensional 3D joint position features. \vspace{-0.05in}
							
\subsection{Experimental Setup}
\label{setup}
\vspace{-0.02in}
%\comment{parameter initialization}
In our experiments, the model parameters $\mathbf{A},
\mathbf{W}, \mathbf{V}$ of the hidden-unit logistic model were
initialized by sampling them from a Gaussian distribution with
a variance of $10^{-3}$. The initial-state parameter
$\boldsymbol{\pi}$, final-state parameter $\boldsymbol{\tau}$
and the bias parameters $\mathbf{b}, \mathbf{c}$ were
initialized to 0. Training of our model is performed using a
standard stochastic gradient descent procedure; the learning
rate is decayed during training. We set the number of hidden
units $H$ to $100$. The L2-regularization parameter $\lambda$
was tuned by minimizing the error on a small validation set.

\begin{figure}[t]
\begin{center}
%\fbox{\rule{0pt}{2in} \rule{0.8\linewidth}{0pt}}
   \includegraphics[width=0.8\linewidth]{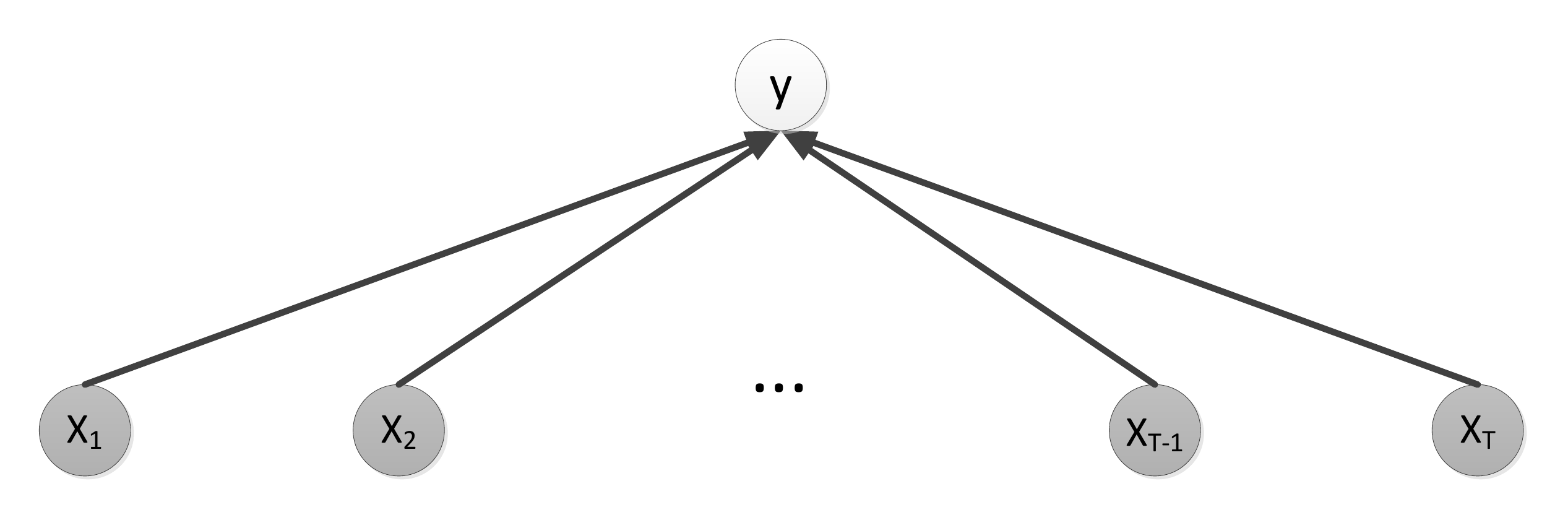}
\end{center}
\vspace{-0.2in}
   \caption{Graphical model of the naive logistic model.}
\label{fig:naive_logistic_model}
\end{figure}

We compare the performance of our hidden-unit logistic model
with that of three other time series classification models: (1)
the naive logistic model shown in
Figure~\ref{fig:naive_logistic_model}, (2) the popular HCRF
model~\cite{HCRF}, and (3) Fisher kernel learning
model~\cite{FKL}. Details of these models are given below.

\paragraph{Naive logistic model.} The naive logistic model is a linear logistic model that shares
parameters between all time steps, and makes a prediction by
summing (or equivalently, averaging) the inner products between
the model weights and feature vectors over time before applying
the softmax function. Specifically, the naive logistic model
defined the following conditional distribution over the label
$y$ given the time series data $\mathbf{x}_{1, \dots, T}$:
\[
p(\mathbf{y} | \mathbf{x}_{1, \dots, T}) = \frac{\mathrm{exp}\{E(\mathbf{x}_{1, \dots, T}, \mathbf{y} )\}}{Z(\mathbf{x}_{1, \dots, T})},
\]
where the energy function is defined as
\[
E(\mathbf{x}_{1, \dots, T}, \mathbf{y} ) = \sum_{t=1}^T {( \mathbf{y}^T \mathbf{W} \mathbf{x}_t}) + \mathbf{c}^T \mathbf{y}.
\]
The corresponding graphical model is shown in
Figure~\ref{fig:naive_logistic_model}. We include the naive
logistic model in our experiments to investigate the effect of
adding hidden units to models that average energy contributions
over time.

%\begin{figure}[t]
%\begin{center}
%%\fbox{\rule{0pt}{2in} \rule{0.9\linewidth}{0pt}}
%\includegraphics[width=0.8\linewidth]{image/naive_logistic_model}
%\end{center}
%   \caption{Factor graph of naive logistic model.  }
%\label{fig:naive_logistic_model}
%\end{figure}

%\comment{simple description about HCRF model and FKL model }
\paragraph{Hidden CRF.} The hidden-state CRF's graphical model is very similar to
that of the hidden-unit logistic model~\cite{HCRF}. The key
difference between the two models is in the way the hidden
units are defined: whereas the hidden-unit logistic model uses
a substantial number of binary stochastic hidden units to
represent the latent state, the HCRF uses a single multinomial
unit (much like a hidden Markov model). We performed
experiments using the hidden CRF implementation
of~\cite{HCRF_code}, which learns the parameters of the model
using L-BFGS. Following~\cite{HCRF}, we trained HCRFs with 10
latent states on all data sets. (We found it was
computationally infeasible to train HCRFs with more than 10
latent states.) We tune the L2-regularization parameter of the
HCRF on a small validation set.

\paragraph{Fisher kernel learning.} In addition to comparing with HCRFs, we compare the performance
of our model with that of the recently proposed Fisher kernel
learning (FKL) model~\cite{FKL}. We selected the FKL model for
our experiments because \cite{FKL} reports strong performance
on a range of time series classification problems. We trained
FKL models based on hidden Markov models with $10$ hidden
states (the number of hidden states was set identical to that
of the hidden CRF). Subsequently, we computed the Fisher kernel
representation and trained a linear SVM on the resulting
features to obtain the final classifier. The slack parameter
$C$ of the SVM is tuned on a small validation set.
\vspace{-0.05in}

\subsection{Results}
\label{result}
\vspace{-0.02in}
We perform two sets of experiments with the hidden-unit
logistic model: (1) a set of experiments in which we evaluate
the performance of the model (and of the hidden CRF) as a
function of the number of hidden units and (2) a set of
experiments in which we compare the performance of all models
on all data sets. The two sets of experiments are described
separately below.
\vspace{-0.1in}
\subsubsection{Effect of Varying the Number of Hidden Units.}
\label{hidden-unit_imapact}
\begin{figure}[t!]
\begin{center}
$\begin{array}{cc}
  \includegraphics[width=0.4\linewidth]{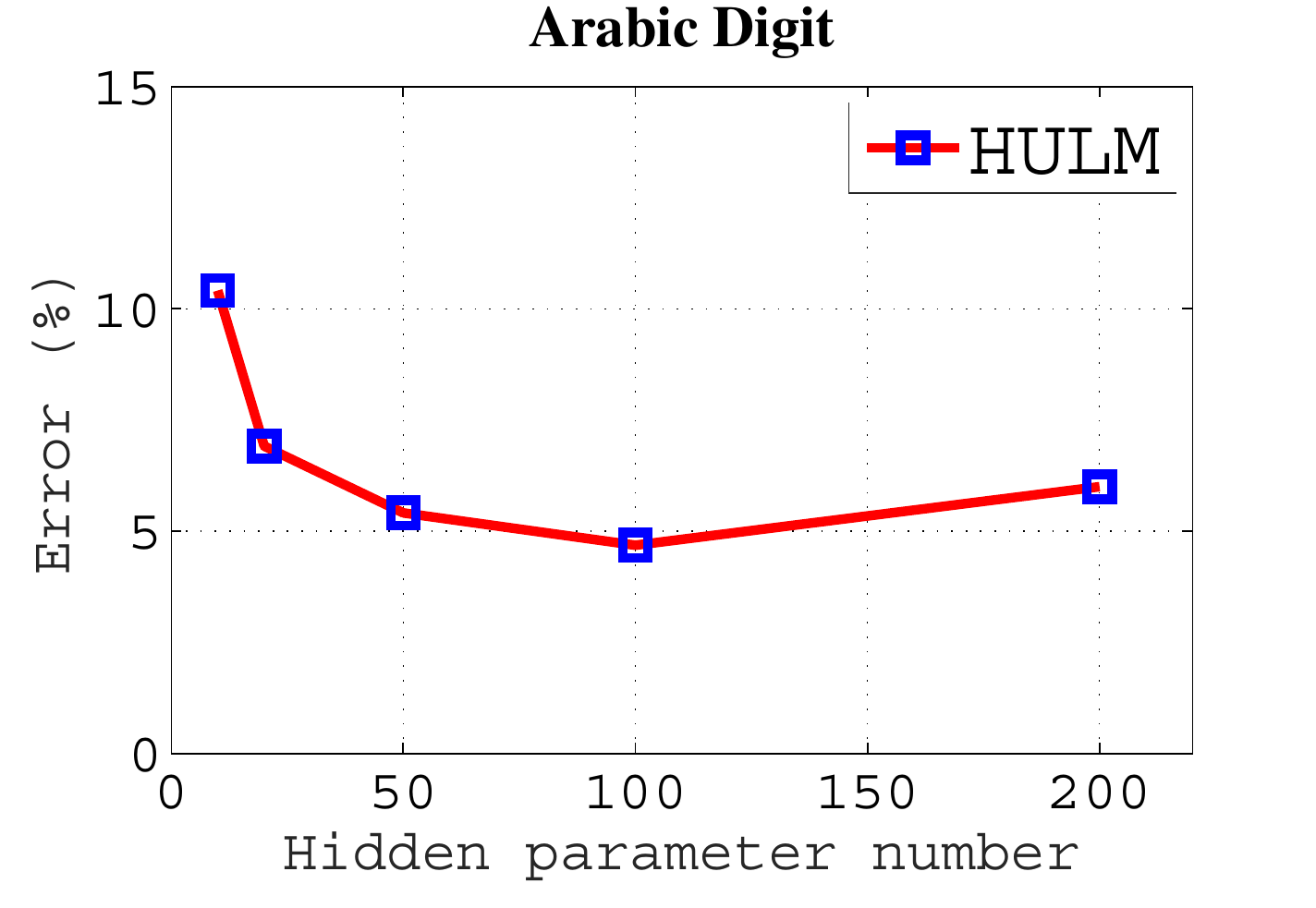} & \includegraphics[width=0.4\linewidth]{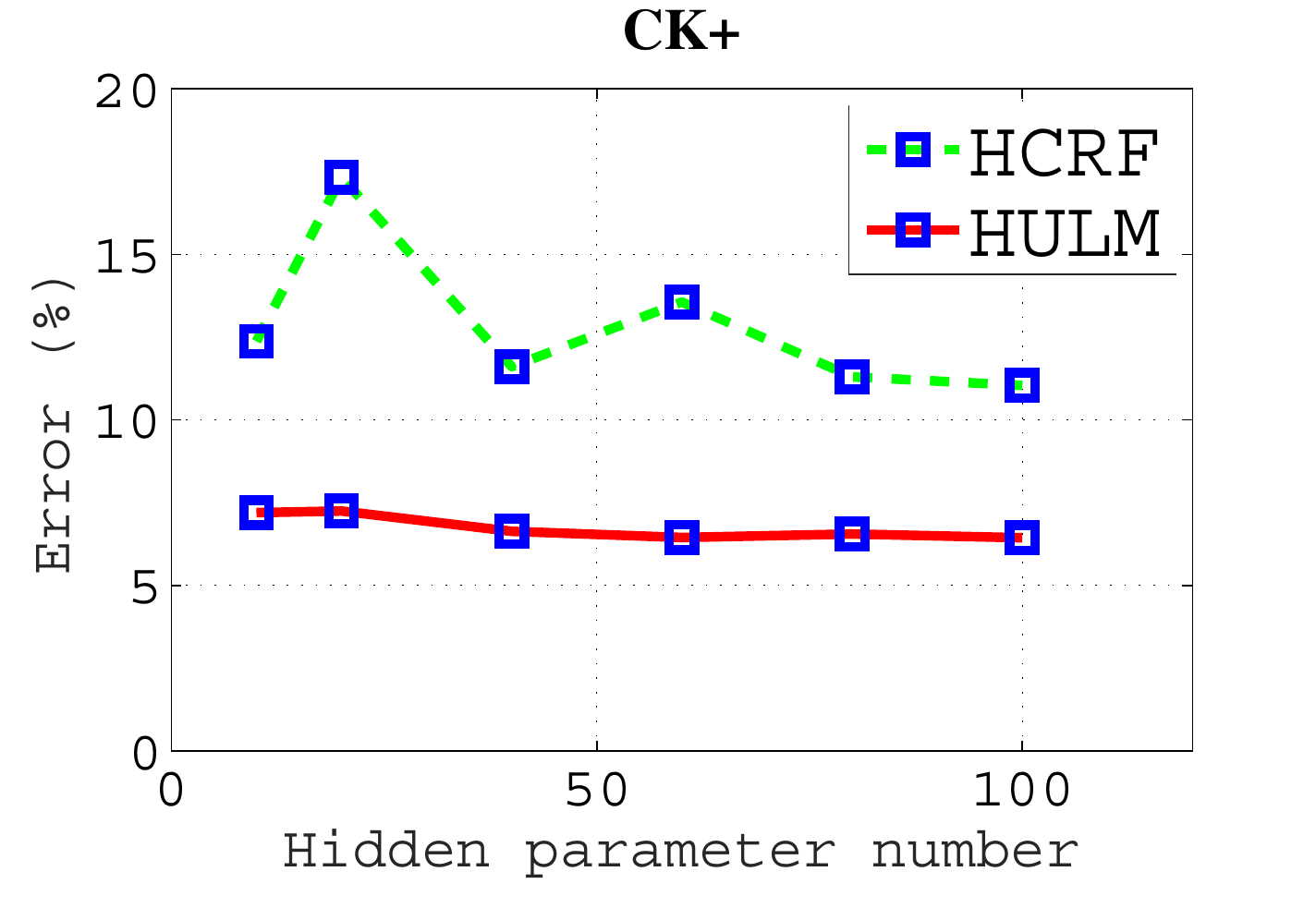} \\
  (a) & (b)   \\
\end{array}$
\end{center}
\vspace{-0.2in}
   \caption{(a) Generalization error ($\%$) of the hidden-unit logistic model on the Arabic speech data set as a function of the number of hidden units. (b) Generalization error ($\%$) of the hidden-unit logistic model and the hidden CRF on the CK+ data set as a function of the hidden parameter number.}
\label{fig:hidden-unit number}
\end{figure}

%\textbf{The impact of hidden-unit number on the performance}.
%\comment{(it hasn't been done yet (in Table~\ref{table:third}).)}
We first conduct experiments on the ASD data set to investigate
the performance of the hidden-unit logistic model as a function
of the number of hidden units. The results of these experiments
are shown in Figure~\ref{fig:hidden-unit number}(a). The
results presented in the figure show that the error initially
decreases when the number of hidden unit increases, because
adding hidden units adds complexity to the model that allows it
to better fit the data. However, as the hidden unit number
increases further, the model starts to overfit on the training
data despite the use of L2-regularization.
%Hence, hidden-unit number $H$ is a very important hyper-parameter affecting the performance of our model. It achieves the best performance at the point of 100 hidden-units. Unless otherwise indicated, we specified the hidden-unit number to be 100 in the left experiments.

We performed a similar experiment on the CK+ facial expression
data set, in which we also performed comparisons with the
hidden CRF for a range of values for the number of hidden
states. Figure~\ref{fig:hidden-unit number}(b) presents the
results of these experiments. On the CK+ data set, there are no
large fluctuations in the errors of the HULM as the hidden
parameter number increases. The figure also shows that the
hidden-unit logistic model outperforms the hidden CRF
irrespective of the number of hidden units. For instance, a
hidden-unit logistic model with 10 hidden units outperforms
even a hidden CRF with 100 hidden parameters. This result
illustrates the potential merits of using models in which the
number of latent states grows exponentially with the number of
parameters.

\vspace{-0.1in}
\subsubsection{Comparison with Modern Time Series Classifiers.}
In a second set of experiments, we compare the performance of
the hidden-unit logistic model with that of the naive logistic
model, Fisher kernel learning, and the hidden CRF on all five
data sets. In our experiments, the number of hidden units in
the hidden-unit logistic model was set to 100;
following~\cite{HCRF}, the hidden CRF used 10 latent states.
The results of our experiments are presented in
Table~\ref{table:first}, and are discussed for each data set
separately below.

\paragraph{Online handwritten character dataset (OHC).} Following the experimental setup in \cite{FKL}, we measure the
generalization error of all our models on the online
handwritten character dataset using 10-fold cross validation.
The average generalization error of each model is shown in
Table~\ref{table:first}. Whilst the naive logistic model
performs very poorly on this data set, all three other methods
achieve very low error rates. The best performance is obtained
FKL, but the differences between the models are very small on
this data set, presumably, due to a ceiling effect.

\begin{table}[t!]
\caption{Generalization errors ($\%$) on all five data sets by
four time series classification models: the naive logistic
model (NL), Fisher kernel learning (FKL), the hidden CRF
(HCRF), and the hidden-unit logistic model (HULM). The best
performance on each data set is boldfaced. See text for
details.}
\begin{center}
\scriptsize
\renewcommand\arraystretch{1.2}
%\resizebox{0.6\linewidth}{!}{
\begin{tabular}{L{2cm}C{1.2cm}C{1.2cm}C{1cm}C{1cm}C{1cm}C{1cm}}
%\Xhline{1pt}
\hline

\multirow{2}{*}{Dataset} & \multirow{2}{*}{Dim.} &
\multirow{2}{*}{Classes} & \multicolumn{4}{c}{Model}
\\ \cline{4-7}
 & & &  NL & FKL & HCRF & HULM\\ %[0.5ex]
\hline
OHC & 3\!$\times$\!10 & 20   & 23.67 &\textbf{0.97}   & 1.58 & 1.30     \\
ASD  & 13\!$\times$\!3 & 10& 25.50   &6.91  & \textbf{3.68} & 4.68     \\
CK+    & 136& 7 & 9.20    &10.81    &11.04  & \textbf{6.44}     \\
Action  & 570 & 20& 40.40  &  40.74   &   \textbf{34.68}  &  35.69\\
Activity  & 570& 16& 59.38   &  \textbf{43.13} & 62.50    &  45.63    \\
[0.2ex] \hline
Avg. error  & -- & -- &  31.63 & 20.51 & 22.70 & \textbf{18.75} \\
Avg. rank  & -- & -- &  3.2 & 2.4 & 2.6 & \textbf{1.8} \\
%[0.5ex]
\hline
%\Xhline{1pt}
\end{tabular}%}
\end{center}
\label{table:first}
\vspace{-0.2in}
\end{table}

\paragraph{Arabic spoken digits dataset (ASD).} Following~\cite{Arabic}, the error rates for the Arabic spoken
digits data set in Table~\ref{table:first} were measured using
a fixed training/test division: $75\%$ of samples are used for
training and left $25\%$ of samples compose test set. The best
performance on this data set is obtained by the hidden CRF
model ($3.68\%$), whilst our model has a slightly higher error
of $4.68\%$, which in turn is better than the performance of
FKL. It should be noted that the performance of the hidden CRF
and the hidden-unit logistic model are better than the error
rate of $6.88\%$ reported in~\cite{Arabic} (on the same
training/test division).

\paragraph{Facial expression dataset (CK+).} Table~\ref{table:first} presents generalization errors measured
using 10-fold cross-validation. Folds are constructed in such a
way that all videos by the same subject are in the same fold
(the subjects appearing in test videos were not present in the
training set). On the CK+ data set, the hidden-unit logistic
model substantially outperforms the hidden CRF model, obtaining
an error of $6.44\%$. Somewhat surprisingly, the naive logistic
model also outperforms the hidden CRF model with an error of
$9.20\%$. A possible explanation for this result is that the
classifying these data successfully does not require
exploitation of temporal structure: many of the expressions can
also be recognized well from a single frame. As a result, the
naive logistic model may perform well even though it simply
averages over time. This result also suggests that the hidden
CRF model may perform poorly on high-dimensional data (the CK+
data is 136-dimensional) despite performing well on
low-dimensional data such as the handwritten character data set
(3-dimensional) and the Arabic spoken data set
(13-dimensional).

\paragraph{MSR Action 3D data set (Action).} To measure the
generalization error of the time series classification models
on the MSR Action 3D dataset, we followed the experimental
setup of~\cite{MSRActivity}: we used all videos of the five
subjects for training, and used the videos of the remaining
five subjects for testing. Table~\ref{table:first} presents the
average generalization error on the videos of the five test
subjects. The four models perform quite similarly, although the
hidden CRF and the hidden-unit logistic model do appear to
outperform the other two models somewhat.

\paragraph{MSR Daily Activity 3D data set (Activity).} On the MSR Daily Activity data set, we use the same
experimental setup as on the action data set: five subjects are
used for training and five for testing. The results in
Table~\ref{table:first} show that the hidden-unit logistic
model substantially outperforms the hidden CRF on this
challenging data set (but FKL performs slightly better).
%the accuracy of our model is around $59\%$ while the reported accuracy of SVM on Fourier Temporal Pyramid (FTP) with only joint position feature in~\cite{MSRActivity} is $68\%$. However, our reproduced experimental result is $65\%$, the difference may be caused by the different  hyper-parameter tuning. The 3D joint positions extracted by the skeleton tracker are very noisy since the performers stand close to the sofa or sit on the sofa. Thus this dataset is very challenging.

In terms of the average error rate and average rank over all
data sets, the hidden-unit logistic model performs very
strongly. Indeed, it substantially outperforms the hidden CRF
model, which illustrates that using a collection of
(conditionally independent) hidden units may be a more
effective way to represent latent states than a single
multinomial unit. FKL also performs quite well in our
experiments, although its performance is slightly worse than
that of the hidden-unit logistic model. However, it should be
noted here that FKL scales poorly to large data sets: its
computational complexity is quadratic in the number of time
series, which limits its applicability to relatively small data
sets (with fewer than, say, $10,000$ time series). By contrast,
the training of hidden-unit logistic models scales linearly in
the number of time series and, moreover, can be performed using
stochastic gradient descent.

%------------------------------------------------------------------------
\section{Application to Facial AU Detection}
In this section, we present a system for facial action unit
(AU) detection that is based on the hidden-unit logistic model.
We evaluate our system on the Cohn-Kanade extended facial
expression database (CK+)~\cite{CK}, evaluating its ability to
detect 10 prominent facial action units: namely, AU1, AU2, AU4,
AU5, AU6, AU7, AU12, AU15, AU17, and AU25. We compare the
performance of our facial action unit detection system with
that of state-of-the-art systems for this problem. Before
describing the results of these experiments, we first describe
the feature extraction of our AU detection system and the setup
of our experiments.

%The flow of the system can be given as follows. Initially, 68
%facial feature points are located/tracked in videos using
%active appearance models (AAM) as described
%in~\cite{maaten2012action}. Using the tracked points, faces are
%normalized to eliminate rigid transformations such as
%translation, rotation, and scale of faces. After normalization,
%shape and appearance features are extracted from each frame.
%Resulting feature sequences are modeled using HULM individually
%for each AU.

\subsection{Facial Features}
\vspace{-0.02in}
We extract two types of features from the video frames in the
CK+ data set: (1) shape features and (2) appearance features.
Our features are identical to the features used by the system
described in~\cite{maaten2012action}; the features are publicly
available online. For completeness, we briefly describe both
types of features below.

The \emph{shape features} represent each frame by the
vertical/horizontal displacements of facial landmarks with
respect to the first frame. To this end, automatically
detected/tracked 68 landmarks are used to form 136-dimensional
time series. All landmark displacements are normalized
by removing rigid transformations (translation, rotation, and scale).

The \emph{appearance features} are based on grayscale
intensity values. To capture the change in facial appearance,
face images are warped onto a base shape, where feature points
are in the same location for each face. After this shape
normalization procedure, the grayscale intensity values of the
warped faces can be readily compared. The final appearance
features are extracted by subtracting the warped textures from
the warped texture in the first frame. The dimensionality of
the appearance feature vectors is reduced using principal
components analysis as to retain $90\%$ of the variance in the
data. This leads to 439-dimensional appearance feature vectors,
which are combined with the shape features to form the final
feature representation for the video frames. For further
details on the feature extraction, we refer to~\cite{maaten2012action}.
\vspace{-0.1in}
%All frames of the videos are normalized with respect to a base
%shape to remove rigid transformations. The rigid
%transformations can be described as changes in translation,
%rotation, and scale of faces during the videos. After
%normalization, features are extracted from each frame. Finally,
%the feature sequences are modeled using HULM.

%In our system, two types of features are extracted from the
%normalized faces, namely, shape and appearance features.
%Movements of facial feature points are good descriptors to
%distinguish between different facial actions. Therefore, in
%order to measure non-rigid facial movements, we subtract the
%normalized landmark coordinates of the first frame from the
%normalized coordinates of the latter frames. Resulting features
%will be referred to as shape features in the remainder of this
%paper.

\subsection{Experimental Setup}
\vspace{-0.02in}
To gauge the effectiveness of the hidden-unit logistic model in
facial AU detection, we performed experiments on the CK+
database~\cite{CK}. The database consists of 593 image
sequences (videos) from 123 subjects with an average length of
18.1 frames. The videos show expressions from neutral face to
peak formation, and include annotations for 30 action units. In
our experiments, we only consider the 10 most frequent action
units.

Our AU detection system employs 10 separate binary classifiers
for detecting action units in the videos. In other words, we
train a separate HULM for each facial
action unit. An individual model thus distinguishes between the
presence and non-presence of the corresponding action unit. We
use a 10-fold cross-validation scheme to measure the
performance of the resulting AU detection system: we randomly
select one test fold containing 10\% of the videos, and use
remaining nine folds are used to train the system. The folds
are constructed such that there is no subject overlap
between folds: \emph{i.e.}, subjects appearing in the test
data were not present in the training data.

\subsection{Results}
\vspace{-0.02in}

We ran experiments using the HULM on three feature sets: (1)
shape features, (2) appearance features, and (3) a
concatenation of both feature vectors. We measure the
performance of our system using the area under ROC curve (AUC).
Table~\ref{table:auc} shows the results for HULM, and for the
baseline in~\cite{maaten2012action}. The results show that the
HULM outperforms the CRF baseline of~\cite{maaten2012action},
with our best model achieving an AUC that is approximately
$0.03$ higher than the best result of~\cite{maaten2012action}.
%\vspace{-0.2in}

%Detailed performance analysis of HULM, using combined features,
%is shown in~\ref{table:comparsion}, where accuracy (ACC),
%recall (RC), precision (PR), F1, AUC measures, and number of
%positive samples are given for each AU.
%\begin{table}[!h]
%\caption{Performance of HULM for different AUs using combined
%features. P shows the number of positive samples. ACC, RC, and
%denote detection accuracy, recall, and precision, respectively.
%\label{table:performance}}
%\begin{center}
%\footnotesize
%%\renewcommand\arraystretch{1.4}
%\begin{tabular}{lcccccc}
%\hline
%AU	&P &ACC	&RC	    &PR	    &F1	    &AUC \\
%\hline
%1	&175 &0.95	&0.88	&0.93	&0.91	&0.96\\
%2	&117 &0.94	&0.84	&0.86	&0.85	&0.96\\
%4	&194 &0.86	&0.71	&0.83	&0.76	&0.90\\
%5	&102 &0.88	&0.62	&0.64	&0.63	&0.88\\
%6	&123 &0.88	&0.63	&0.77	&0.69	&0.92\\
%7	&121 &0.82	&0.58	&0.56	&0.57	&0.81\\
%12	&131 &0.95	&0.88	&0.89	&0.88	&0.95\\
%15	&95  &0.91	&0.75	&0.70	&0.72	&0.92\\
%17	&203 &0.92	&0.91	&0.87	&0.89	&0.97\\
%25	&324 &0.95	&0.95	&0.97	&0.96	&0.97\\
%\hline
%Avg.&-&0.91	&0.77	&0.80	&0.79	&0.93\\
%\hline
%\end{tabular}
%\end{center}
%\end{table}
\begin{table}[!t]
\caption{AUC of the HULM and the CRF baseline
in~\cite{maaten2012action} for three feature sets.
*In~\cite{maaten2012action}, the combined feature set also
includes SIFT features.\label{table:auc}}
\begin{center}
\scriptsize
\renewcommand\arraystretch{1.2}
\begin{tabular}{L{1.5cm}C{1.5cm}C{1.5cm}C{1.7cm}}
\hline
\multirow{2}{*}{Method} & \multicolumn{3}{c}{Feature Set}      \\ \cline{2-4}
                        &Shape   &Appearance & Combination    \\ \hline
HULM                        &0.9101  &0.9197      &0.9253 \\
\cite{maaten2012action}     &0.8902  &0.8971      &0.8647* \\
\hline
\end{tabular}
\end{center}
\end{table}
\begin{figure}[!t]
\begin{center}
\begin{tabular}{cC{1cm}c}
\includegraphics[width=0.25\linewidth]{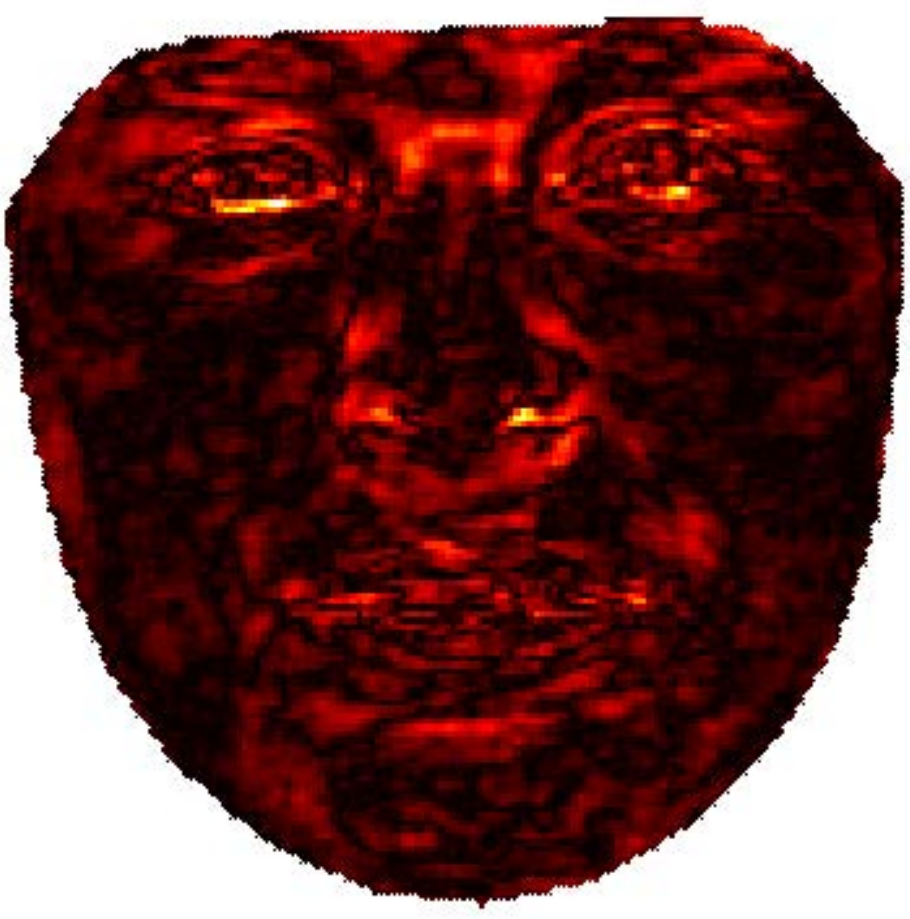}   & &
\includegraphics[width=0.25\linewidth]{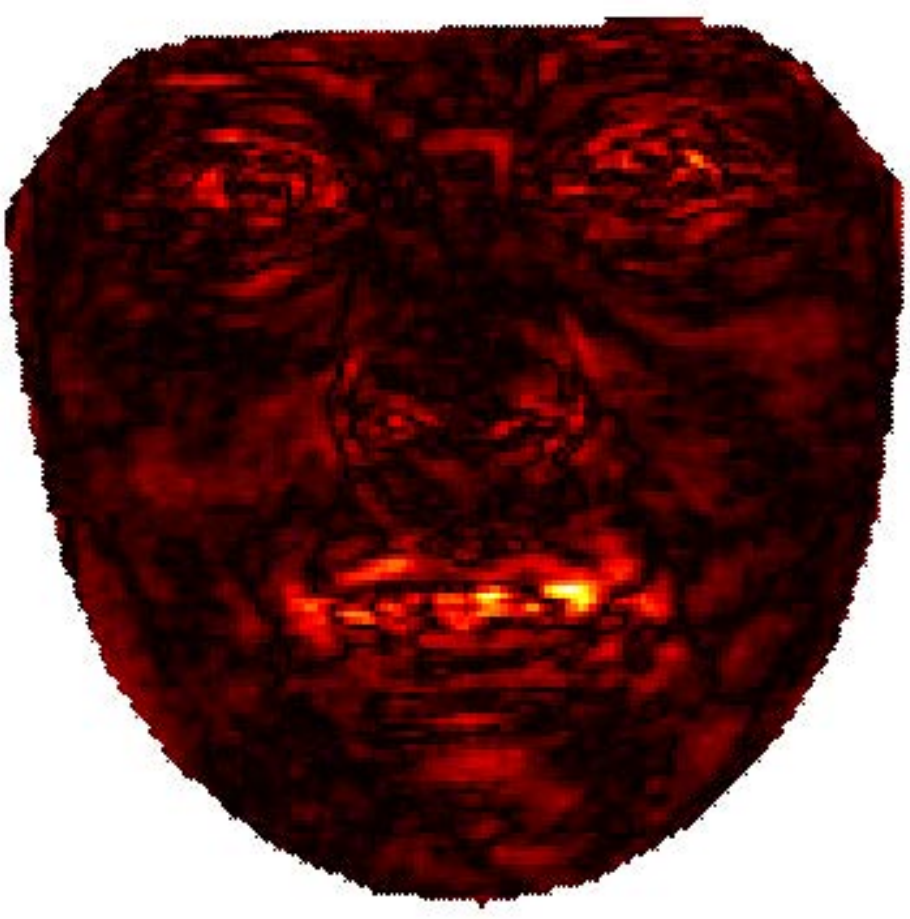}  \\
 (a) && (b) \\
\end{tabular}
\end{center}
\vspace{-0.25in}
\caption{Visualization of $|\mathbf{W}|$ for (a) AU4 and (b) AU25. Brighter colors correspond to image regions with higher weights. \label{fig:auVis}}
\end{figure}

To obtain insight in what features are modeled by the HULM
hidden units, we visualized a single column of $|\mathbf{W}|$
in Figure~\ref{fig:auVis} for the AU4 and AU25 models that were
trained on appearance features. Specifically, we selected the
hidden unit with the highest corresponding $\mathbf{V}$-value
for visualization, as this hidden unit apparently models the
most discriminative features. The figure shows that the
appearance of the eyebrows is most important in the AU4 model
(brow lowerer), whereas the mouth region is most important in
the AU25 model (lips part).
%High
%displays on the other regions (which are not related) can be
%caused by the co-occurring AUs with AU4 and AU25. Note that
%colormaps are individually scaled for Figure~\ref{fig:auVis}(a)
%and Figure~\ref{fig:auVis}(b) to able to amplify the relative
%differences.

\begin{table}[!t]
\caption{Average F1-scores of our system and seven state-of-the-art systems on the CK+ data set. The F1
scores for all methods were obtained from the literature. Note that the averages are not over the same AUs, and cannot readily be compared. The best performance for each condition is boldfaced.
\label{table:comparsion}}
\begin{center}
\scriptsize
\renewcommand\arraystretch{1.1}
%\resizebox{1.0\linewidth}{!}{
\begin{tabular}{C{1cm}C{1cm}C{1cm}C{1cm}C{1cm}C{1cm}C{1cm}C{1cm}C{1cm}} \hline
AU	&HULM	&\cite{koelstra2010dynamic}	
&\cite{valstar2012fully}	&\cite{li2013data}	
&\cite{li2013TIP}
&\cite{ding2013COT}	&\cite{zhang2014p}\\
\hline
1 &\textbf{0.91}	&0.87	&0.83	&0.66	&0.78	&0.76	&0.88\\
2 &0.85	&0.90	&0.83	&0.57	&0.80	&0.76	&\textbf{0.92}\\
4 &0.76	&0.73	&0.63	&0.71	&0.77	&0.79	&\textbf{0.89}\\
5 &0.63	&\textbf{0.80}	&0.60	&--	    &0.64	&--	    &--\\
6 &0.69	&0.80	&0.80	&\textbf{0.94}	&0.77	&0.70	&0.93\\
7 &0.57	&0.47	&0.29	&\textbf{0.87}	&0.62	&0.63	&-- \\
12&0.88	&0.84	&0.84	&0.88	&\textbf{0.90}	&0.87	&\textbf{0.90}\\
15&0.72	&0.70	&0.36	&\textbf{0.84}	&0.70	&0.71	&0.73\\
17&\textbf{0.89}	&0.76	&--	    &0.79	&0.81	&0.86	&0.76\\
25&\textbf{0.96}	&\textbf{0.96}	&0.75	&--	    &0.88	&--	    &0.73\\
\hline
Avg. & 0.79  &0.78	        &0.66	&0.78	&0.77	&0.76	&0.84\\\hline
%Avg. of HULM& 0.79 &\textbf{0.79}	&\textbf{0.77}	&\textbf{0.78}	&\textbf{0.79}	&\textbf{0.78}	&0.83\\
\end{tabular} %}
\end{center}
\end{table}
%\vspace{-0.1in}

In Table~\ref{table:comparsion}, we compare the performance of
our AU detection system with that of seven other
state-of-the-art systems in terms of the more commonly used
F1-score. (Please note that the averages are not over the same
AUs, and cannot readily be compared.) The results in the table
show that our system achieves the best F1 scores for AU1, AU17,
and AU25. It performs very strongly on most of the other AUs,
illustrating the potential of the hidden-unit logistic model.

\begin{comment}
Detailed performance analysis of the proposed hidden-unit
logistic model (HULM), using combined features, is given in
Table~\ref{table:performance}, where accuracy (ACC), recall
(RC), precision (PR), F1, AUC measures, and number of positive
samples are given for each AU.
\begin{table}[htb]
\caption{Performance of HULM for different AUs using combined
features. P shows the number of positive samples. ACC, RC, and
denote detection accuracy, recall, and precision, respectively.
\label{table:performance}}
\begin{center}
\footnotesize
%\renewcommand\arraystretch{1.4}
\begin{tabular}{lcccccc}
\hline
AU	&P &ACC	&RC	    &PR	    &F1	    &AUC \\
\hline
1	&175 &0.95	&0.88	&0.93	&0.91	&0.96\\
2	&117 &0.94	&0.84	&0.86	&0.85	&0.96\\
4	&194 &0.86	&0.71	&0.83	&0.76	&0.90\\
5	&102 &0.88	&0.62	&0.64	&0.63	&0.88\\
6	&123 &0.88	&0.63	&0.77	&0.69	&0.92\\
7	&121 &0.82	&0.58	&0.56	&0.57	&0.81\\
12	&131 &0.95	&0.88	&0.89	&0.88	&0.95\\
15	&95  &0.91	&0.75	&0.70	&0.72	&0.92\\
17	&203 &0.92	&0.91	&0.87	&0.89	&0.97\\
25	&324 &0.95	&0.95	&0.97	&0.96	&0.97\\
\hline
Avg.&-&0.91	&0.77	&0.80	&0.79	&0.93\\
\hline
\end{tabular}
\end{center}
\end{table}
\vspace{50cm}.
%Note that the state-of-the-art methods used in this
%comparison have specifically designed and optimized for AU
%detection task, while our approach is a direct application of
%the proposed hidden-unit logistic model.
\end{comment}

%------------------------------------------------------------------------
\section{Conclusions}
In this paper, we presented the hidden-unit logistic model
(HULM), a new model for the single-label classification of time
series. The model is similar in structure to the popular hidden
CRF model, but it employs binary stochastic hidden units
instead of multinomial hidden units between the data and label.
As a result, the HULM can model exponentially more latent
states than a hidden CRF with the same number of parameters.
The results of our experiments with HULM on several real-world
datasets show that this may result in improved performance on
challenging time-series classification tasks. In particular,
the HULM performs very competitively on complex computer-vision
problems such as facial expression recognition.

%LAURENS: This should go somewhere in the results section

%It is shown in Section~\ref{result} that FKL performs on par with our model. However, one of key disadvantages of FKL model is that FKL scales quadratically in the number of time series, which makes it infeasible to dataset with big number of time series(eg. $> 5000$). 	

In future work, we aim to explore more complex variants of our hidden-unit logistic model. In particular, we intend to study variants of the model in which the simple first-order Markov chains on the hidden units are replaced by more powerful, higher-order temporal connections. Specifically, we intend to implement the higher-order chains via a similar factorization as used in neural autoregressive distribution estimators \cite{larochelle11}. The resulting models will likely have longer temporal memory than our current model, which will likely lead to stronger performance on complex time series classification tasks. A second direction for future work we intend to explore is an extension of our model to multi-task learning. Specifically, we will explore multi-task learning scenarios in which sequence labeling and time series classification is performed simultaneously (for instance, simultaneous recognition of short-term actions and long-term activities, or simultaneous optical character recognition and word classification). By performing sequence labeling and time series classification based on the same latent features, the performance on both tasks may be improved because information is shared in the latent features.

\section*{Acknowledgments}
This work was supported by EU-FP7 INSIDDE and AAL SALIG++.

\begin{small}
\bibliographystyle{splncs03}
\bibliography{egbib}
\end{small}

\end{document}